\documentclass{l4dc2027}

\usepackage{times}
\usepackage{mathtools}
\usepackage{booktabs}
\usepackage{float}
\usepackage{wrapfig}
\usepackage{enumitem}
\usepackage{etoolbox}

\makeatletter
\patchcmd{\@jmlrenddoc}{\null}{}{}{}
\makeatother

\title[GP--RL--MPC]{Uncertainty-Aware Reinforcement Learning-Augmented Model Predictive Control with Gaussian Process Postdecision State-Value Functions}

\author{%
  \Name{Jon Li} \Email{zhiangli@usc.edu}
  \\
  \Name{Jay H. Lee} \Email{jlee4140@usc.edu}
  \AND
  \addr Mork Family Department of Chemical Engineering and Materials Science, University of Southern California, Los Angeles, CA 90089, United States
}

\begin{document}
\maketitle

\begin{abstract}
We propose Gaussian Process Reinforcement Learning-Augmented Model Predictive Control (GP--RL--MPC), which uses a learned critic as the MPC terminal penalty to account for costs beyond the prediction horizon. Gaussian process regression (GPR) enables data-efficient critic learning from limited, noisy rollouts within optimistic approximate policy iteration (OAPI), with soft updates mitigating policy chattering. A tunable risk coefficient weights GP epistemic uncertainty in the terminal cost, regulating sensitivity to uncertain critic predictions. The postdecision state formulation shifts disturbance averaging offline to account for aleatoric uncertainty from the terminal transition onward. Under matching objectives and admissible actions, a one-step horizon recovers greedy value-space RL. After fitting GP hyperparameters by exact marginal likelihood, we construct a sparse deterministic training conditional (DTC) posterior using inducing points selected by pivoted Cholesky (PC), reducing online evaluation costs as the training dataset grows. In a continuous stirred-tank reactor (CSTR) subject to exogenous disturbances, GP--RL--MPC with an appropriately selected risk coefficient and policy screening reduces the sum of squared errors (SSE) by an average of $80.52$\% relative to an otherwise matched baseline MPC across five random seeds (sample standard deviation: $3.48$\%).
\end{abstract}


\section{Introduction}
\label{sec:introduction}

Model predictive control (MPC) handles nonlinear dynamics, multivariable interactions, and operating constraints through repeated finite-horizon optimization~\citep{Lee2026}. Computational demands and growing prediction uncertainty limit practical horizon lengths, making terminal costs important for accounting for long-term consequences beyond the prediction horizon. Under appropriate assumptions, suitable terminal ingredients can support recursive feasibility, closed-loop stability, and performance guarantees. However, constructing such terminal ingredients remains challenging, particularly for nonlinear stochastic systems.

Reinforcement learning (RL) offers a complementary approach by learning long-term cost information from closed-loop rollouts to approximate terminal values in MPC~\citep{Bertsekas2024}. Previous studies have explored approximate dynamic programming (ADP) for control~\citep{Lee2006}, RL-based parameterization of economic nonlinear MPC (NMPC)~\citep{Gros2020}, and terminal-value learning using approximate value iteration~\citep{MorenoMora2023}, partially input-convex neural networks (PICNNs)~\citep{Li2026a}, and radial basis function (RBF) networks~\citep{Li2026b}. A key challenge in extending these approaches to stochastic systems is incorporating the effects of random state transitions into the learned continuation value without substantially increasing the computational burden of online MPC optimization.

The conventional (predecision) state-value function represents the expected cumulative cost conditioned on the current state, but evaluating it at a nominal terminal state omits averaging over disturbed terminal states. Explicitly computing this additional expectation online adds to the burden of stochastic MPC~\citep{Mesbah2016}. A postdecision state captures the known effect of the control action before the stochastic transition, allowing its value function to incorporate the expectation over successor states~\citep{Sutton2018,Powell2016}. Learning this function from disturbed rollouts shifts disturbance averaging offline while preserving deterministic online MPC optimization. A one-step horizon recovers greedy value-space RL, while longer horizons provide multistep lookahead.

Building on the postdecision state formulation, we develop GP--RL--MPC to improve short-horizon MPC performance while preserving its computational advantages. A Gaussian process (GP) critic learns the conditional expected cost-to-go from disturbed rollouts under a fixed policy, thereby incorporating future disturbance effects into postdecision value estimates evaluated at nominal MPC terminal states. The terminal penalty is defined as $\mu_{\mathrm{GP}}^p+\kappa\sigma_{\mathrm{GP}}^p$, where $\mu_{\mathrm{GP}}^p$ and $\sigma_{\mathrm{GP}}^p$ denote the GP posterior mean and latent posterior standard deviation, respectively. The tunable risk coefficient $\kappa$ regulates the influence of critic epistemic uncertainty on control decisions~\citep{Rasmussen2006}.

Critic learning is integrated into an optimistic approximate policy iteration (OAPI) framework, with soft updates mitigating policy chattering~\citep{Bertsekas2019}. GP hyperparameters are estimated by maximizing the exact marginal likelihood, after which pivoted Cholesky (PC) selects inducing points for a sparse deterministic training conditional (DTC) approximation~\citep{Harbrecht2012,QuinoneroCandela2005}. This sparse representation makes online posterior mean and variance evaluation costs depend primarily on the inducing-set size rather than the full training-set size, reducing computational costs while retaining estimates of critic epistemic uncertainty.

\begin{figure}[!t]
    \centering
    \includegraphics[width=\linewidth]{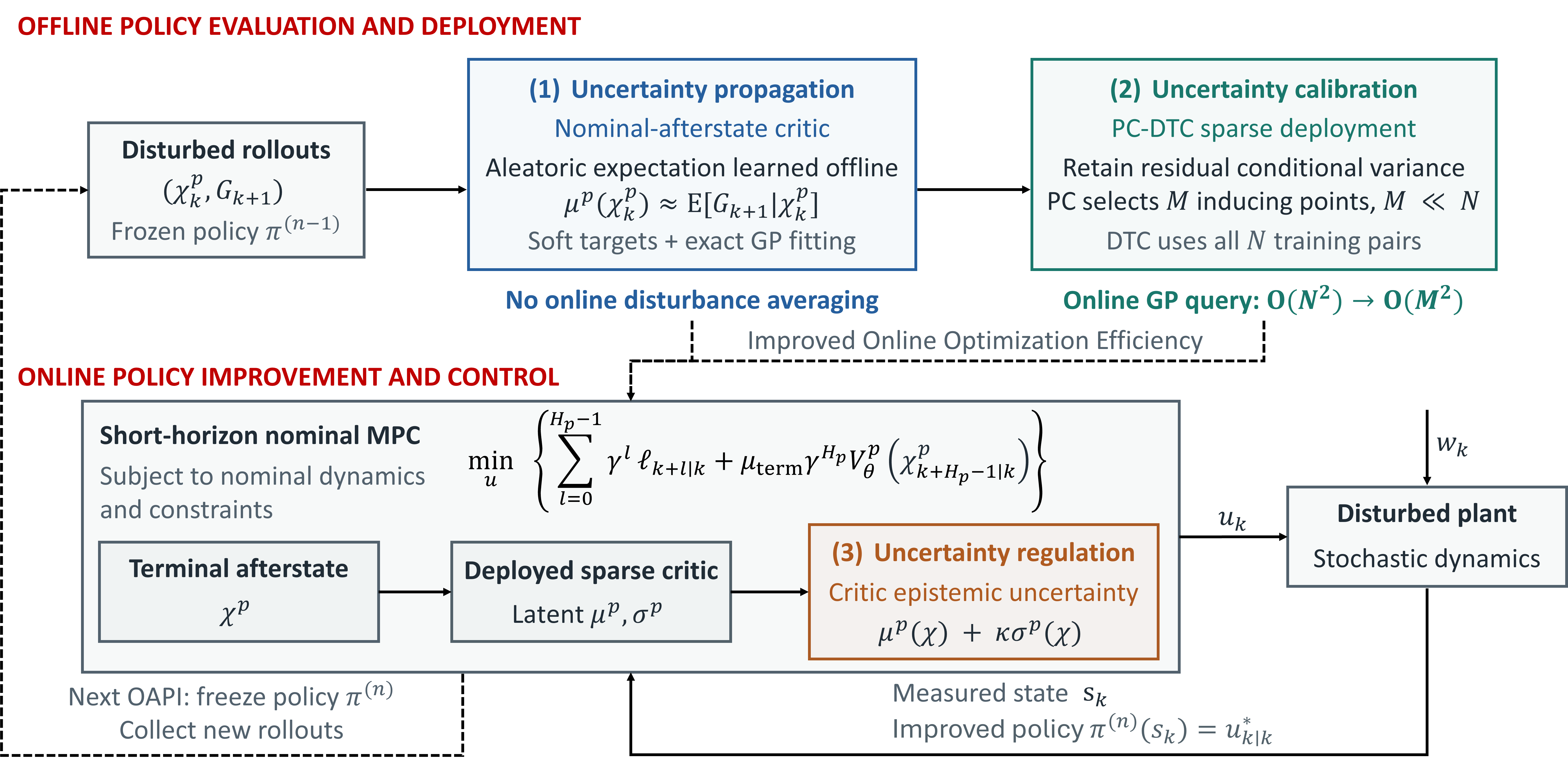}
    \caption{Overview of the proposed GP--RL--MPC algorithm.}
    \label{fig:gprlmpc-algorithm}
    \vspace{-12pt}
\end{figure}

The main contributions of the proposed uncertainty-aware framework, illustrated in Fig.~\ref{fig:gprlmpc-algorithm}, are as follows:
\begin{itemize}[nosep]
\item \textbf{Uncertainty propagation:} A postdecision state-value critic that learns the expected effects of aleatoric uncertainty on the cost-to-go from disturbed rollouts, enabling nominal MPC to account for future disturbances without explicitly evaluating expectations during online optimization.
\item \textbf{Uncertainty calibration:} A computationally efficient GP critic that combines exact marginal-likelihood hyperparameter estimation with pivoted Cholesky (PC) inducing-point selection and deterministic training conditional (DTC) sparsification, reducing online computational costs while retaining estimates of epistemic uncertainty.
\item \textbf{Uncertainty regulation:} A terminal penalty that incorporates GP epistemic uncertainty through a tunable risk coefficient, combined with soft critic updates within optimistic approximate policy iteration, to promote reliable policy improvement while preserving the efficiency of short-horizon MPC.
\end{itemize}

\section{Preliminaries}
\label{sec:preliminary}

\subsection{Problem Formulation}
\label{ssec:problem_formulation}

For nominal prediction, consider the discrete-time nonlinear model
\begin{equation}
    \label{eq:discrete-model}
    x_{k+1}=f_s(x_k,d_k,u_k),
\end{equation}
where $x_k \in\mathbb{R}^{n_x}$, $d_k \in\mathbb{R}^{n_d}$, and $u_k \in\mathbb{R}^{n_u}$ denote the system state, information state, and manipulated input, respectively. The information state collects references, time-varying parameters, disturbances, and information related to model-plant mismatch. The controlled output is given by $y_k=h(x_k)$, where $h:\mathbb{R}^{n_x}\rightarrow\mathbb{R}^{n_y}$, assuming no direct input feedthrough. The stochastic components of the information state are modeled as zero-drift random walks, $d_{k+1}=d_k+\varepsilon_k,$ where $\{\varepsilon_k\}_{k\ge 0}$ is a sequence of independent and identically distributed (i.i.d.) Gaussian innovations with \(\varepsilon_k\sim\mathcal{N}(0,R_{\varepsilon})\). Known references and exogenous forecasts evolve according to prescribed trajectories.

Given $x_k$ and $d_k$, or their estimates, the nominal NMPC problem is formulated as~\citep{Lee2011,Li2025}
\begin{subequations}
    \label{eq:nmpc}
    \begin{align}
        J_{H_p}(x_k)
        \coloneqq
        \min_{\{u_{k+l\mid k}\}_{l=0}^{H_p-1}}
        &
        \begin{aligned}[t]
            \sum_{l=0}^{H_p-1}
            \gamma^l
            \ell&(x_{k+l\mid k}, d_{k+l\mid k},u_{k+l\mid k})
            +
            \gamma^{H_p}
            \Psi(x_{k+H_p\mid k},d_{k+H_p\mid k}),
        \end{aligned}
        \\
        \text{s.t.}\quad
        &
        x_{k\mid k}=x_k,\qquad d_{k\mid k}=d_k,
        \\
        &
        x_{k+l+1\mid k}
        =
        f_s(x_{k+l\mid k},d_{k+l\mid k},u_{k+l\mid k}),
        \\
        &
        d_{k+l+1\mid k}=d_{k+l\mid k},
        \\
        &
        x_{k+l\mid k}\in\mathcal{X},\qquad
        u_{k+l\mid k}\in\mathcal{U},
        \\
        &
        x_{k+H_p\mid k}\in\mathcal{X}_f.
    \end{align}
\end{subequations}
Here, $x_{k+l\mid k}$, $d_{k+l\mid k}$, and $u_{k+l\mid k}$ denote the predicted state, information state, and manipulated input, respectively, at $l$ steps ahead, as computed at time $k$, and $H_p$ denotes the prediction horizon. Setting future stochastic innovations to zero yields certainty-equivalent (CE) predictions without explicitly propagating disturbance uncertainty. The admissible sets, $\mathcal{X}$ and $\mathcal{U}$, are assumed to be closed and compact, and the stage cost $\ell(x,d,u)$ quantifies the tracking or economic performance~\citep{Rawlings2020}. While conventional deterministic MPC typically employs an undiscounted objective, we introduce a discount factor $\gamma\in(0,1)$ to ensure consistency with infinite-horizon value learning under persistent stochastic disturbances, assuming that the expected discounted cumulative cost is finite.

In conventional MPC, a terminal set $\mathcal{X}_f\subseteq\mathcal{X}$ and penalty $\Psi$ can ensure recursive feasibility and closed-loop stability under suitable assumptions but are challenging to construct for nonlinear stochastic systems. We learn the terminal cost from closed-loop rollouts to capture long-term information while retaining a short prediction horizon for computational efficiency. In this study, the proposed controller enforces hard state and input constraints within the prediction horizon.  The numerical results shown in Section~\ref{sec:experiments} demonstrate that the states of interest remain within their prescribed bounds with substantial safety margins~\citep{Li2026b}. Selected state constraints can be softened by penalizing bounded violations while retaining hard input constraints and outer state bounds. Using the same augmented stage cost in the MPC objective and Monte Carlo (MC) returns from disturbed rollouts incorporates expected future violation costs into the learned postdecision state-value function. No additional terminal violation penalty is required; Appendix~\ref{app:soft-constraints} provides the detailed formulation and its consistency with the value-learning targets.

\subsection{Postdecision State-Value Functions}
\label{ssec:postdecision}

A postdecision state separates the known effects of a control action from the subsequent stochastic transition~\citep{Wong2011}. When the system dynamics admit such a decomposition, the state transition can be expressed as
\begin{equation}
    s_k^p=F_{\mathrm a}(s_k,u_k), \qquad s_{k+1}=F_{\mathrm w}(s_k^p,w_k).
\label{eq:postdecision-transition}
\end{equation}
Here, $s_k\in\mathcal{S}$ denotes the augmented predecision state, comprising the plant state $x_k$, information state $d_k$, and any additional variables, such as input memory or contextual information, required to satisfy the Markov property. The superscript $p$ denotes the postdecision state, and $w_k$ represents the stochastic innovation realized after the control action. For the postdecision state to provide an exact representation, it must contain sufficient information to characterize the conditional distribution of the next predecision state, such that
\begin{equation}
    \mathbb{P}(s_{k+1}\mid s_k,u_k)=\mathbb{P}(s_{k+1}\mid s_k^p).
    \label{eq:postdecision-sufficiency}
\end{equation}

For a fixed policy $\pi$, the postdecision state-value function is defined as
\begin{equation}
V_{\pi}^p(s_k^p)\coloneqq\mathbb{E}\!\left[V_{\pi}(s_{k+1})\mid s_k^p\right]=\mathbb{E}_{\pi}\!\left[G_{k+1}\mid s_k^p\right],
\label{eq:postdecision-value}
\end{equation}
where $V_{\pi}$ is the conventional predecision state-value function and $G_{k+1}\coloneqq\sum_{j=k+1}^{\infty}\gamma^{j-k-1}\ell(s_j,u_j)$ denotes the Monte Carlo (MC) return starting at time $k+1$, with subsequent actions determined by $\pi$. Thus, the postdecision state-value function captures the expected cumulative cost arising from the next stochastic transition and all subsequent disturbances, while excluding the current stage cost. Under the sufficiency condition, the Bellman expectation equation for a fixed deterministic policy $\pi$ becomes
\begin{equation}
V_{\pi}(s)=\ell\bigl(s,\pi(s)\bigr)+\gamma V_{\pi}^p\bigl(F_{\mathrm a}(s,\pi(s))\bigr).
\label{eq:postdecision-bellman}
\end{equation}
The corresponding one-step policy improvement is obtained by minimizing the right-hand side of Eq.~\eqref{eq:postdecision-bellman} with respect to the current control input. Importantly, because the postdecision state-value function already incorporates the expectation over subsequent stochastic transitions, this minimization does not require explicit disturbance averaging. This property makes the postdecision formulation particularly attractive for integration with nominal MPC, allowing stochastic effects on future costs to be accounted for without increasing the complexity of online optimization. Further details on the construction of postdecision training targets and their consistency with nominal MPC terminal-state queries are provided in Appendix~\ref{app:postdecision-conventions}.

\subsection{Postdecision State-Value Learning as Optimistic Approximate Policy Iteration}
\label{ssec:oapi}

Starting from a baseline MPC policy $\pi^{(0)}$ without a terminal penalty, each policy iteration (PI)~$\#n$ evaluates the preceding policy $\pi^{(n-1)}$ using disturbed rollouts. The resulting data are used to update the GP critic, whose posterior mean $\mu_{\mathrm{GP}}^{p,(n)}$ approximates the postdecision state-value function $V_{\pi^{(n-1)}}^p$. The GP posterior mean and latent posterior standard deviation are then combined to construct the uncertainty-regulated terminal value estimate $V_\theta^{p,(n)}$, which is incorporated into the constrained MPC optimization to obtain a candidate policy $\pi^{(n)}$. Thus, the critic learning serves as approximate policy evaluation, while the MPC optimization performs policy improvement using the learned postdecision state-value function~\citep{Bertsekas2024}. This iterative procedure constitutes optimistic approximate policy iteration (OAPI)~\citep{Bertsekas2022}, in which policy improvement is performed following partial, sample-based policy evaluation rather than waiting for complete convergence of the value function. Such incomplete evaluation can lead to oscillations between successive policies, commonly referred to as policy chattering~\citep{Li2026a,Li2026b}. The conventional predecision state-value function and its relationship to the postdecision state-value function are reviewed in Appendix~\ref{app:standard-value-function}.

\section{Proposed GP--RL--MPC Algorithm}
\label{sec:algorithm}

The proposed GP--RL--MPC algorithm integrates approximate policy evaluation and policy improvement by iteratively learning a postdecision state-value critic from disturbed closed-loop rollouts and updating the MPC policy through constrained nominal optimization. At PI~$\#n$, the preceding policy $\pi^{(n-1)}$ is held fixed while disturbed rollouts are generated to obtain MC training pairs $(\chi_k^{p,(r)},G_{k+1}^{(r)})$, where $r$ indexes rollouts and $\chi_k^{p,(r)}\coloneqq\varphi(s_k^{p,(r)})$ denotes the feature representation of the postdecision state used as input to the GP critic. These training pairs are pooled and reindexed to form the dataset $\mathcal D_{n-1}^p=\{(\chi_i^p,G_i)\}_{i=1}^{N_{\mathrm{pool}}}$ for GP regression. The construction of the training pairs and their temporal alignment with the postdecision states are detailed in Appendix~\ref{app:postdecision-conventions}. The resulting GP posterior mean $\mu_{\mathrm{GP}}^{p,(n)}$ and latent posterior standard deviation $\sigma_{\mathrm{GP}}^{p,(n)}$ are used to construct an uncertainty-regulated terminal penalty for the MPC optimization, yielding an updated control policy. Soft critic updates are employed to mitigate policy chattering across successive iterations. The complete GP--RL--MPC procedure is summarized in Algorithm~\ref{alg:gp-rl-mpc}.

\subsection{Soft Updates in Value Space}
\label{ssec:soft-update}

At the first policy iteration, the GP critic is trained directly on the Monte Carlo returns $G_i$. From the second iteration onward, soft updates are introduced by constructing the regression targets as
\begin{equation}
    y_i^{(n)}
    =(1-\tau)\mu_{\mathrm{GP}}^{p,(n-1)}(\chi_i^p)+\tau G_i,
    \qquad 0<\tau\leq1,
\label{eq:value-space-soft-update}
\end{equation}
where $\tau$ is the soft-update parameter and $\mu_{\mathrm{GP}}^{p,(n-1)}$ is the posterior mean of the preceding GP critic, held fixed during target construction. The update is performed in the original cost units, blending the previous critic prediction with the newly observed MC return. Smaller values of $\tau$ provide stronger damping of critic updates, thereby mitigating abrupt changes in the learned value function and the resulting MPC policy. Setting $\tau=1$ recovers the raw MC targets, corresponding to a hard update.

The resulting input-target pairs are partitioned into training and diagnostic subsets using target-quantile stratification~\citep{Li2026b}. Input and output scalers are fitted using the initial training subset and subsequently held fixed to ensure consistent scaling across policy iterations. At each iteration, the GP hyperparameters are initialized from their values in the preceding iteration and refitted using the current training dataset. This procedure combines damped policy evaluation with consistent data preprocessing and warm-started GP training to improve the stability of successive policy updates.

\subsection{Exact GPR Training and PC-DTC Deployment}
\label{ssec:pc-dtc}

We fit the full GP critic offline by maximizing the exact log marginal likelihood. With the fitted hyperparameters fixed, pivoted Cholesky (PC) selects $M\ll N$ inducing points from the $N$ training inputs~\citep{Harbrecht2012}. A deterministic training conditional (DTC) approximation incorporates information from all training pairs through this inducing representation~\citep{QuinoneroCandela2005}. The latent uncertainty excludes observation noise and represents critic epistemic uncertainty conditional on the fitted GP model. After precomputing the Cholesky factors and predictive mean coefficients offline, evaluating the mean and variance at one query requires $\mathcal O(M^2+Mn_\chi)$ operations, compared with $\mathcal O(N^2+Nn_\chi)$ for exact GPR using dense Cholesky factors, where $n_\chi$ is the input dimension~\citep{Rasmussen2006}. The regression model, inducing-point selection, and predictive formulas are detailed in Appendix~\ref{app:gpr}.

\subsection{Uncertainty-Regulated Terminal Cost}
\label{ssec:risk-terminal}

Using the deployed GP posterior moments, we construct an uncertainty-regulated terminal cost based on an upper confidence bound (UCB) acquisition rule:
\begin{equation}
    V_\theta^p(\chi)
    \coloneqq\mu_{\mathrm{GP}}^p(\chi)
    +\kappa\sigma_{\mathrm{GP}}^p(\chi),
\label{eq:risk-terminal}
\end{equation}
where $\kappa\in\mathbb R$ controls the influence of epistemic uncertainty. Positive $\kappa$ penalizes uncertain value estimates, yielding a more conservative controller; $\kappa=0$ uses the posterior mean alone, and negative $\kappa$ favors actions leading to uncertain regions, thereby encouraging exploration. Incorporating this approximation at the postdecision terminal feature $\chi_{k+H_p-1\mid k}^p$ gives the GP--RL--MPC objective
\begin{equation}
    J_{H_p}^{\mathrm{GP}}(s_k;\theta)
    \coloneqq\min_{\{u_{k+l\mid k}\}_{l=0}^{H_p-1}}
    \Bigg\{\sum_{l=0}^{H_p-1}\gamma^l\ell(s_{k+l\mid k},u_{k+l\mid k})
    +\mu_{\mathrm{term}}\gamma^{H_p}
    V_\theta^p
    \bigl(\chi_{k+H_p-1\mid k}^p\bigr)\Bigg\},
\label{eq:gprlmpc-objective}
\end{equation}
subject to the dynamics, disturbance recursion, initialization, and state and input constraints in Eq.~\eqref{eq:nmpc}. The first optimal input defines the receding-horizon policy $\pi^{(n)}(s_k)\coloneqq u_{k\mid k}^{\star,(n)}$. After applying this input, the optimization is repeated at the next sampling instant. The terminal weight $\mu_{\mathrm{term}}>0$ adjusts the contribution of the learned value function, with $\mu_{\mathrm{term}}=1$ corresponding to its unscaled contribution. Setting the terminal approximation to zero recovers the baseline MPC policy $\pi^{(0)}$.

The stage cost penalizes tracking errors and input changes:
\begin{equation}
    \ell(s_k,u_k)=\|y_k-r_k\|_{\mu_y}^2+\|u_k-u_{k-1}\|_{\mu_u}^2,
\label{eq:stage-cost}
\end{equation}
where $\mu_y,\mu_u\ge0$ are the tracking and input-move weights.

\subsection{OAPI Procedure and Policy Assessment}
\label{ssec:algorithm}

At each optimistic approximate policy iteration (OAPI), the updated critic defines a new MPC policy, which is evaluated through closed-loop rollouts. For tracking applications, policy performance is measured by the mean rollout sum of squared errors (SSE):
\begin{equation}
    S_n=\operatorname{Score}(\pi^{(n)})
    \coloneqq
    \frac{1}{R}\sum_{r=1}^{R}\sum_{k=0}^{T-1}
    \left\|y_k^{(r)}-r_k\right\|_2^2,
    \label{eq:policy-score}
\end{equation}
where $R$ is the number of rollouts collected under each fixed policy, and $T$ is the number of sampling intervals per rollout. Unlike the MPC objective, this evaluation metric is undiscounted and excludes input-move penalties, providing a direct measure of tracking performance. The same rollouts provide both the training batch for the next iteration and the score used for policy screening. 

Pseudocode~\ref{alg:gp-rl-mpc} summarizes the OAPI procedure. The routine $\operatorname{Collect}(\pi)$ returns the postdecision training dataset and mean SSE obtained under policy $\pi$, following the rollout protocol in Appendix~\ref{app:postdecision-conventions}. At each iteration, the candidate policy is compared with the best accepted policy. A candidate is rejected if its mean SSE increases; otherwise, it is accepted, and iteration terminates when the percentage improvement falls below $\delta_{\mathrm{PI}}$ or the maximum number of iterations $N_{\max}$ is reached.

\begin{algorithm2e}[!t]
\caption{OAPI for GP--RL--MPC with postdecision state-value learning}
\label{alg:gp-rl-mpc}
\DontPrintSemicolon
\KwIn{Baseline $\pi^{(0)}$; shared rollout settings; maximum PI iterations $N_{\max}$; improvement threshold $\delta_{\mathrm{PI}}$; MPC and GPR settings.}
\KwOut{Selected controller $\pi_{\mathrm{best}}$.}

$(\mathcal D_0^p,S_0)\leftarrow\operatorname{Collect}(\pi^{(0)})$\;
$n^\star\leftarrow0$\;

\For{$n=1,\ldots,N_{\max}$}{
    \lIf{$S_{n^\star}=0$}{\textbf{break}}
    Construct targets from $\mathcal D_{n-1}^p$: raw returns at $n=1$; otherwise Eq.~\eqref{eq:value-space-soft-update}\;
    Split data using target-quantile stratification\;
    Standardize using scalers fitted on the initial training subset and held fixed thereafter\;
    Fit $\theta^{(n)}$ by maximizing the exact GP marginal likelihood, warm-starting for $n\ge2$\;
    Select PC inducing points and construct the DTC posterior using Eqs.~\eqref{eq:dtc-mean} and~\eqref{eq:dtc-var}\;
    Form $V_{\theta^{(n)}}^p$ and $\pi^{(n)}$ using Eqs.~\eqref{eq:risk-terminal} and~\eqref{eq:gprlmpc-objective}\;
    $(\mathcal D_n^p,S_n)\leftarrow\operatorname{Collect}(\pi^{(n)})$\;
    Store the candidate policy, critic, and score\;
    
    $\Delta_n\leftarrow100(S_{n^\star}-S_n)/S_{n^\star}$\;
    \eIf{$\Delta_n<0$}{
    Reject $\pi^{(n)}$ and retain $\pi^{(n^\star)}$ and its critic\;
    \textbf{break}\;
    }{
    Accept candidate $\pi^{(n)}$ and its critic; set $n^\star\leftarrow n$\;
    \lIf{$\Delta_n<\delta_{\mathrm{PI}}$}{\textbf{break}}
    }
}

\Return{$\pi_{\mathrm{best}}=\pi^{(n^\star)}$}\;
\end{algorithm2e}

\section{Experiments}
\label{sec:experiments}

\subsection{Van de Vusse CSTR}
\label{ssec:cstr}

We evaluate GP--RL--MPC on a Van de Vusse continuous stirred-tank reactor (CSTR), a classical process systems engineering (PSE) benchmark~\citep{Li2026a}. The reaction network consists of consecutive reactions $A\rightarrow B\rightarrow C$, which produce and subsequently consume the desired intermediate $B$, and a parallel reaction $2A\rightarrow D$, which consumes reactant $A$. Within each sampling interval, the plant dynamics are 
\begin{subequations}
\label{eq:cstr}
\begin{align}
    \dot x_1
    &=-k_1x_1-k_3x_1^2+(x_{1f}-x_1)u_k,
    \label{eq:cstr-x1}\\
    \dot x_2
    &=k_1x_1-k_2x_2-x_2u_k
    +\omega_k^{x_2},
    \label{eq:cstr-x2}\\
    \dot x_{1f}
    &=\omega_k^{x_{1f}},
    \label{eq:cstr-feed}
\end{align}
\end{subequations}
where $x_1$ and $x_2$ denote the concentrations of $A$ and $B$, respectively, $x_{1f}$ is the feed concentration, and $u_k$ is the dilution rate. The kinetic coefficients are $k_1,k_2,$ and $k_3$. The stochastic forcing terms satisfy $\omega_k^{x_2}\sim\mathcal N(0,\sigma_{x_2}^2)$ and $\omega_k^{x_{1f}}\sim\mathcal N(0,\sigma_{x_{1f}}^2)$, independently across channels and sampling intervals. Each disturbance is sampled once per interval and applied as a constant forcing rate under a zero-order hold (ZOH). Consequently, the feed concentration evolves as a random walk, while nominal MPC predictions assume zero future innovations. The plant parameters, controller settings, and operating conditions are summarized in Table~\ref{app-tab:cstr-parameters} of Appendix~\ref{app:experiment-configurations}.

The controlled output is the concentration of the desired product, $y_k=x_{2,k}$, with reference $r_k$ and stage cost defined by Eq.~\eqref{eq:stage-cost}. The feed concentration is included in the information state as $d_k=x_{1f,k}$. The nominal postdecision state feature vector is 
\begin{equation}
    \chi_k^p
    \coloneqq\varphi(s_k^p)
    =
    \begin{bmatrix}
        \bar x_{1f,k+1\mid k}\\
        r_{k+1\mid k}\\
        \bar x_{1,k+1\mid k}\\
        \bar x_{2,k+1\mid k}-r_{k+1\mid k}
    \end{bmatrix},
    \label{eq:cstr-postdecision-features}
\end{equation}
where overbars denote nominal endpoints obtained from the current state and applied input with zero disturbance innovations. The GP employs a quadratic mean function, $m_\beta$, in the tracking-error coordinate, together with an automatic-relevance-determination (ARD) Mat\'ern kernel defined over all four features.

The experiments compare GP--RL--MPC with baseline MPC, assess tracking improvements across accepted policy iterations, and examine how the risk coefficient $\kappa$ affects policy-iteration performance.

\subsection{Baseline Comparison and Policy Iteration Performance}
\label{ssec:baseline-pi-performance}

\begin{table}[!t]
    \centering
    \small
    \setlength{\tabcolsep}{6pt}
    \caption{Final incumbent GP--RL--MPC performance with $\kappa=3.0$.}
    \label{tab:screened-pi-kappa3}
    \begin{tabular}{@{}lcccc@{}}
        \toprule
        Seed & Accepted PI & MPC SSE $S_0$ & Final SSE $S_{n^\star}$ & SSE Reduction (\%) \\
        \midrule
        $1$ & $7$ & $3.20$ & $0.46$ & $85.63$ \\
        $2$ & $5$ & $2.30$ & $0.40$ & $82.43$ \\
        $3$ & $6$ & $2.18$ & $0.47$ & $78.38$ \\
        $4$ & $8$ & $1.81$ & $0.42$ & $77.07$ \\
        $5$ & $5$ & $2.49$ & $0.52$ & $79.07$ \\
        \midrule
        Mean $\pm$ SD & --- & --- & --- & $80.52\pm3.48$ \\
        \bottomrule
    \end{tabular}
\end{table}

We evaluate the final incumbent GP--RL--MPC controllers with $\kappa=3.0$ against a baseline MPC controller with zero terminal cost. Both use the same prediction horizon, transition model, stage cost, and constraints. Initial conditions, reference schedules, and disturbance realizations are matched across controllers and PI rounds, and each SSE score is averaged over $R=5$ rollouts. Table~\ref{tab:screened-pi-kappa3} summarizes the final incumbent performance after five to eight accepted PI rounds, during which SSE decreases monotonically. The reductions relative to baseline MPC, calculated as $100(1-S_{n^\star}/S_0)$, range from $77.07\%$ to $85.63\%$ with a mean of $80.52\%$ and a sample standard deviation of $3.48\%$. The consistently large improvements across all five seeds demonstrate the effectiveness of the learned terminal cost in compensating for the limited prediction horizon. These results also provide empirical support for reliable policy improvement within the OAPI framework. The corresponding closed-loop tracking responses are shown in Fig.~\ref{fig:mpc-comparison} in Appendix~\ref{app:cl-results}.

\subsection{Effect of the Risk Coefficient}
\label{ssec:risk-coefficient}

\begin{figure}[!t]
    \centering
    \includegraphics[width=\linewidth]{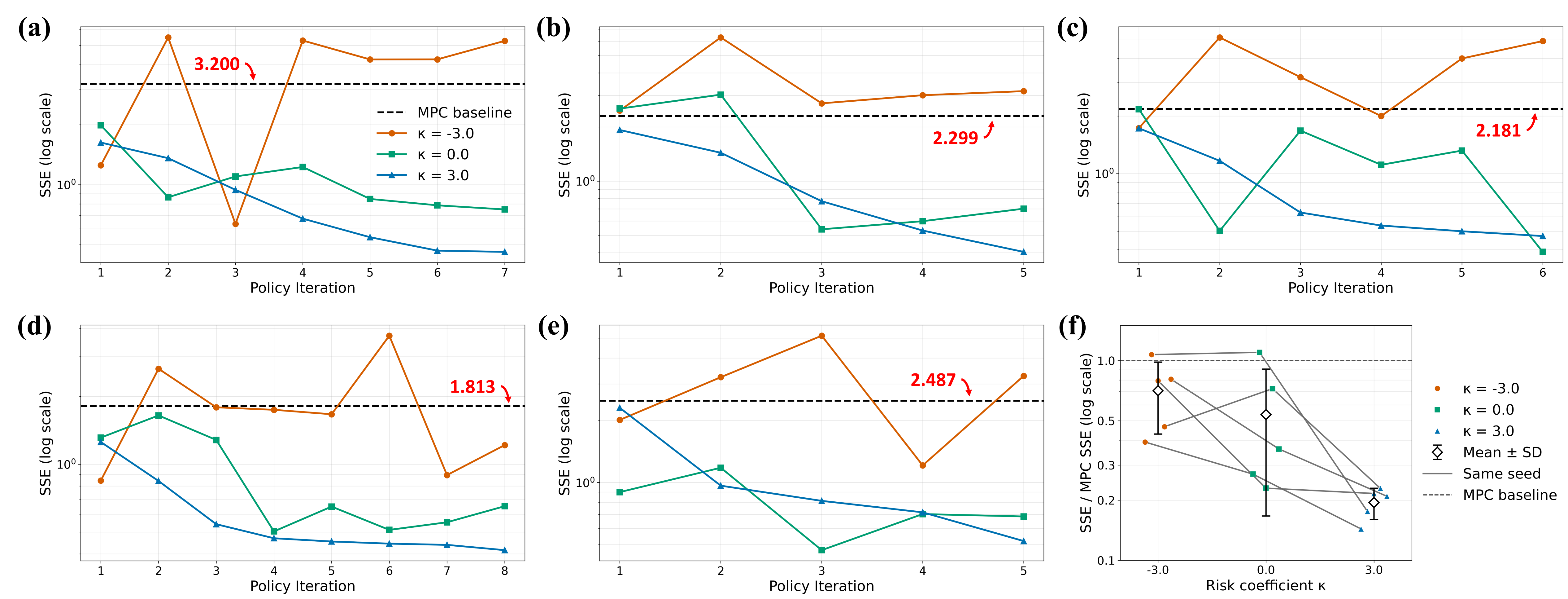}
    \caption{Multi-seed analysis of the risk coefficient: (a)--(e): Mean rollout SSE across policy iterations. (f): Normalized SSE at the end of each initial non-increasing segment.}
    \label{fig:risk-coefficient}
    \vspace{-12pt}
\end{figure}

To investigate the role of epistemic uncertainty in terminal-cost optimization, we compare three risk coefficients, $\kappa\in\{-3.0,0.0,3.0\}$. These correspond to an optimistic lower confidence bound, the posterior mean, and a conservative upper confidence bound, respectively, with $\kappa=\pm3.0$ representing the bounds of a nominal $99.7\%$ pointwise Gaussian credible interval. Fig.~\ref{fig:risk-coefficient}(a)--(e) shows the resulting mean rollout SSE trajectories across policy iterations for the five seed configurations.

The screening-enabled configuration with $\kappa=3.0$ achieves sustained, monotonically non-increasing SSE across accepted PI rounds. In contrast, the $\kappa=0$ and $\kappa=-3.0$ configurations exhibit non-monotonic behavior, with generally larger fluctuations under $\kappa=-3.0$. These latter two configurations are evaluated as diagnostic runs without screening, each continued for the same number of PI rounds as its corresponding $\kappa=3.0$ run. Although mean-based optimization ($\kappa=0.0$) and optimistic optimization ($\kappa=-3.0$) occasionally achieve lower SSE, these improvements are neither predictable nor consistently sustained. For example, in Fig.~\ref{fig:risk-coefficient}(c), $\kappa=0$ achieves an SSE of $0.39$ at PI~$\#6$, compared with $0.47$ for $\kappa=3.0$, but only after several intermediate performance reversals. Such behavior reflects the sensitivity of policy improvement to the accuracy of the learned critic, which depends on the disturbance realizations and the coverage of the relevant state-action space by the GP training data. In particular, optimistic terminal-cost estimates may favor decisions in poorly sampled regions where critic predictions are less reliable.

To compare the consistency of improvement across risk coefficients, Fig.~\ref{fig:risk-coefficient}(f) summarizes the initial non-increasing segments of the SSE trajectories, starting at PI~$\#1$. The score at the end of each segment is normalized by its corresponding baseline score $S_0$. The resulting means and sample standard deviations of normalized SSE are $0.19\pm0.03$, $0.54\pm0.37$, and $0.71\pm0.28$ for $\kappa=3.0$, $0.0$, and $-3.0$, respectively. Under this segment-based comparison, $\kappa=3.0$ achieves the lowest mean normalized SSE and the smallest variability across seeds.

\section{Conclusion}
\label{sec:conclusion}

We developed GP--RL--MPC, an uncertainty-aware framework integrating uncertainty propagation, calibration, and regulation to improve short-horizon MPC performance while preserving its computational advantages. For uncertainty propagation, the GP postdecision state-value critic approximates expected cost-to-go under aleatoric uncertainty from disturbed MC rollouts aligned with nominal terminal queries. This formulation shifts disturbance averaging offline and avoids explicit online expectation calculations. For uncertainty calibration, exact marginal-likelihood hyperparameter fitting is followed by PC-DTC sparse deployment, reducing kernel redundancy and online evaluation costs while maintaining reliable uncertainty estimates. For uncertainty regulation, the UCB-based terminal cost uses the latent posterior standard deviation to provide tunable regulation of critic epistemic uncertainty in control decisions. Numerical results provide empirical support for consistent and sustained policy improvement under the OAPI procedure. Future work will extend GP--RL--MPC to economic optimization and control.

\section*{Generative AI Use}
Generative AI tools were used to assist with writing. The authors reviewed and verified the AI-assisted content and take responsibility for the accuracy and integrity of the submitted work.

\bibliography{bibliography}

\clearpage
\appendix

\section{Soft-Constraint Violation Penalties}
\label{app:soft-constraints}

As an illustrative example, consider softening only the upper product-concentration bound $x_{2,\max}^{\mathrm s}=2.4\,\mathrm{mol\,L^{-1}}$. For $l=0,\ldots,H_p-1$, replace the corresponding hard upper bound in $\mathcal X$ with
\begin{equation}
\begin{aligned}
    x_{2,k+l\mid k} &\leq x_{2,\max}^{\mathrm s}+\xi_{k+l\mid k},\\
    0 &\leq\xi_{k+l\mid k}\leq\xi_{\max},
\end{aligned}
\label{eq:soft-state-bound}
\end{equation}
where $\xi_{\max}>0$ limits the admissible violation. This retains the hard outer bound $x_{2,k+l\mid k}\leq x_{2,\max}^{\mathrm s}+\xi_{\max}$. All other state and input bounds remain unchanged.

Let $[a]_+\coloneqq\max(a,0)$ and define
\begin{equation}
\begin{aligned}
    p_{\mathrm c}(x) &\coloneqq\rho_{\mathrm c}[x_2-x_{2,\max}^{\mathrm s}]_+, \qquad \rho_{\mathrm c}>0, \\
    \ell_{\mathrm c}(s,u) &\coloneqq\ell(s,u)+p_{\mathrm c}(x),
\end{aligned}
\label{eq:penalized-stage-cost}
\end{equation}
where $\ell$ is the tracking and input-move cost in Eq.~\eqref{eq:stage-cost}. When the slack variable appears only in Eq.~\eqref{eq:soft-state-bound} and its linear penalty, the minimizing value is $\xi_{k+l\mid k}=[x_{2,k+l\mid k}-x_{2,\max}^{\mathrm s}]_+$. Thus, within the admissible outer bound, the slack formulation is equivalent to using $\ell_{\mathrm c}$.

The resulting GP--RL--MPC objective becomes
\begin{equation}
    J_{H_p,\mathrm c}^{\mathrm{GP}}(s_k;\theta)
    \coloneqq\min_{\{u_{k+l\mid k}\}_{l=0}^{H_p-1}}
    \Bigg\{\sum_{l=0}^{H_p-1}\gamma^l\Big[\ell(s_{k+l\mid k},u_{k+l\mid k})+\rho_{\mathrm c}\xi_{k+l\mid k}\Big]+\mu_{\mathrm{term}}\gamma^{H_p}V_\theta^p\bigl(\chi_{k+H_p-1\mid k}^p\bigr)\Bigg\},
\label{eq:soft-gprlmpc-objective}
\end{equation}
subject to the nominal dynamics, initialization, remaining hard bounds, and the soft constraints in Eq.~\eqref{eq:soft-state-bound}. The horizon penalties are evaluated on nominal predictions.

For critic learning, the same augmented stage cost is evaluated along the disturbed closed-loop rollout:
\begin{equation}
\begin{aligned}
    G_{k+1}^{\mathrm c} &\coloneqq\sum_{j=k+1}^{T-1}\gamma^{j-k-1}\ell_{\mathrm c}(s_j,u_j),\\
    G_j^{\mathrm c} &=\ell_{\mathrm c}(s_j,u_j)+\gamma G_{j+1}^{\mathrm c},\qquad G_T^{\mathrm c}=0.
\end{aligned}
\label{eq:penalized-mc-return}
\end{equation}
Following the postdecision state-value formulation, each nominal postdecision state feature $\chi_k^p$ is paired with the disturbed return $G_{k+1}^{\mathrm c}$, preserving the indexing and truncation conventions in Appendix~\ref{app:postdecision-conventions}. Violation penalties are evaluated using the full
realized constraint violation, even when it exceeds $\xi_{\max}$. When this extension is enabled, baseline and learned policies use the same penalized cost $\ell_{\mathrm c}$, and $G_i^{\mathrm c}$ replaces $G_i$ in the MC target construction and value-space soft updates.

Because the learned postdecision value incorporates
the augmented stage costs from $k+H_p$ onward, including constraint-violation penalties, no additional terminal violation penalty is required. Any hard terminal set $\mathcal X_f$, if imposed, remains a separate constraint and is not enforced by the learned terminal value. Moreover, the soft
constraints and violation penalties imposed on
nominal predictions do not guarantee constraint
satisfaction for the disturbed plant.

\section{Gaussian Process Regression and PC-DTC Deployment}
\label{app:gpr}

This appendix summarizes exact Gaussian process regression (GPR) and the PC-DTC approximation deployed in Section~\ref{ssec:pc-dtc}. All inputs and regression targets below are expressed in standardized coordinates, and policy-iteration indices are suppressed unless explicitly shown.

\subsection{Exact Gaussian Process Regression}
\label{app:exact-gpr}

Let $\chi\in\mathbb{R}^{n_\chi}$ denote an input and $y\in\mathbb{R}$ the associated regression target. A GP prior on the latent function $v$ is
\begin{equation}
v\sim\mathcal{GP}(m_\beta,k),
\label{eq:gp-prior}
\end{equation}
where $m_\beta$ is the parameterized mean function and $k$ is the covariance kernel. With Gaussian likelihood noise of variance $\sigma_n^2>0$, the observation model is
\begin{equation}
y_i=v(\chi_i)+\nu_i,\qquad \nu_i\overset{\mathrm{i.i.d.}}{\sim}\mathcal{N}(0,\sigma_n^2).
\label{eq:gp-likelihood}
\end{equation}
For $N$ training pairs, write $X=(\chi_1,\ldots,\chi_N)$ and define
\begin{equation}
\label{eq:gp-training-definitions}
\begin{aligned}
\mathbf{y} &= (y_1,\ldots,y_N)^\top,\\
\mathbf{m} &= \bigl(m_\beta(\chi_1),\ldots,m_\beta(\chi_N)\bigr)^\top,\\
[K]_{ij} &= k(\chi_i,\chi_j),\qquad K_y=K+\sigma_n^2I.
\end{aligned}
\end{equation}

For a query $\chi_*$, let $k(X,\chi_*)=\bigl(k(\chi_1,\chi_*),\ldots,k(\chi_N,\chi_*)\bigr)^\top\in\mathbb{R}^{N\times1}$. Conditional on the fitted parameters, the exact latent posterior mean and variance are~\citep{Rasmussen2006}
\begin{align}
\mu_{\mathrm{GP}}(\chi_*) &= m_\beta(\chi_*)+k(X,\chi_*)^\top K_y^{-1}(\mathbf{y}-\mathbf{m}),
\label{eq:gp-posterior-mean}\\
\sigma_{\mathrm{GP}}^2(\chi_*) &= k(\chi_*,\chi_*)-k(X,\chi_*)^\top K_y^{-1}k(X,\chi_*).
\label{eq:gp-posterior-var}
\end{align}

We use an automatic-relevance-determination (ARD) Mat\'ern-$3/2$ kernel,
\begin{equation}
k(\chi,\chi')=\sigma_f^2\bigl(1+\sqrt{3}\,a(\chi,\chi')\bigr)\exp\bigl(-\sqrt{3}\,a(\chi,\chi')\bigr),
\label{eq:matern32-kernel}
\end{equation}
with scaled distance
\begin{equation}
a(\chi,\chi')=\sqrt{\sum_{j=1}^{n_\chi}\frac{(\chi_j-\chi'_j)^2}{\lambda_j^2}}.
\label{eq:matern-distance}
\end{equation}
The signal variance $\sigma_f^2>0$ sets the covariance scale, while the length scales $\lambda_j>0$ control variation along each input coordinate. Larger length scales correspond to slower variation along the associated coordinates.

The posterior mean approaches $m_\beta(\chi_*)$ as the cross-covariances vanish far from the training inputs. The latent posterior variance represents model-based epistemic uncertainty conditional on the fitted GP assumptions and parameters. It excludes observation-noise variance and does not directly quantify stochastic return variability or plant risk.

\subsection{PC-DTC Deployment}
\label{app:pc-dtc}

For critic learning, let $\mathcal D_N=\{(\chi_i^p,y_i)\}_{i=1}^{N}$ denote the current training subset. The regression targets are constructed as described in Section~\ref{ssec:soft-update} and subsequently standardized. The parameter vector $\theta$, comprising the mean function, kernel, and likelihood parameters, is fitted using Adam to maximize the exact GP log marginal likelihood. The PC-DTC approximation is constructed only after fitting, with these parameters held fixed.

For input collections $U$ and $V$, let $K_{UV}$ denote the kernel matrix with entries $[K_{UV}]_{ij}=k(U_i,V_j)$. For a single input $\chi$, let $k(Z,\chi)$ denote the column vector of covariances between $\chi$ and the inducing points in $Z$. Starting with $Z=\varnothing$, PC sequentially selects the unselected training input having the largest residual conditional prior variance~\citep{Harbrecht2012},
\begin{equation}
    \rho_i(Z)
    =
    k(\chi_i^p,\chi_i^p)
    -k(Z,\chi_i^p)^\top K_{ZZ}^{-1}k(Z,\chi_i^p),
\label{eq:pc-selection}
\end{equation}
where the second term is defined as zero when $Z=\varnothing$. The residual $\rho_i(Z)$ measures the prior variance at $\chi_i^p$ unexplained by the current inducing set. Selection terminates when the inducing-point budget is reached, the total residual variance falls below a prescribed fraction of its initial value, or the maximum residual variance becomes numerically negligible. This yields an inducing set of size $M\le M_{\max}$, typically with $M\ll N$.

DTC uses the selected inducing points to approximate the training covariance while incorporating information from all training pairs~\citep{QuinoneroCandela2005}. Write $X=(\chi_1^p,\ldots,\chi_N^p)$, $\mathbf y=(y_1,\ldots,y_N)^\top$, and $m_\beta(X)=\bigl(m_\beta(\chi_1^p),\ldots,m_\beta(\chi_N^p)\bigr)^\top$. Define
\begin{equation}
    B=K_{ZZ}+\delta_{\mathrm{jit}}I,
    \qquad
    A=K_{ZX}K_{XZ}+\sigma_n^2B,
\label{eq:dtc-matrices}
\end{equation}
where $\delta_{\mathrm{jit}}>0$ provides numerical regularization for the Cholesky factorization. Standard DTC is recovered as $\delta_{\mathrm{jit}}\to0$. For $Z=\{z_1,\ldots,z_M\}$ and a query $\chi_*$, the inducing-point covariance vector is $k(Z,\chi_*)\coloneqq\bigl(k(z_1,\chi_*),\ldots,k(z_M,\chi_*)\bigr)^\top\in\mathbb R^{M\times1}$. The resulting predictive mean and latent variance are
\begin{align}
\mu_{\mathrm{DTC}}(\chi_*)
&=m_\beta(\chi_*)+k(Z,\chi_*)^\top A^{-1}K_{ZX}(\mathbf y-m_\beta(X)),
\label{eq:dtc-mean}\\
\sigma_{\mathrm{DTC}}^2(\chi_*)
&=k(\chi_*,\chi_*)-k(Z,\chi_*)^\top B^{-1}k(Z,\chi_*)
+\sigma_n^2k(Z,\chi_*)^\top A^{-1}k(Z,\chi_*).
\label{eq:dtc-var}
\end{align}
The variance combines a regularized residual conditional variance, represented by the first two terms, with posterior uncertainty in the inducing variables, represented by the third term. No observation-noise variance is added to the latent prediction.

The deployed mean $\mu_{\mathrm{GP}}^{p,(n)}$ and latent posterior standard deviation $\sigma_{\mathrm{GP}}^{p,(n)}$ are obtained by transforming $\mu_{\mathrm{DTC}}$ and the square root of $\sigma_{\mathrm{DTC}}^2$ back to the original cost units. These deployed moments replace the exact posterior moments in Eqs.~\eqref{eq:gp-posterior-mean} and~\eqref{eq:gp-posterior-var} when constructing the terminal penalty in Eq.~\eqref{eq:risk-terminal}.

\section{Predecision and Postdecision State-Value Functions}
\label{app:standard-value-function}

Let $s\in\mathcal{S}$ denote the augmented Markov state and $\ell(s,u)$ the stage cost. For a fixed stationary policy $\pi:\mathcal{S}\rightarrow\mathcal{U}$, the standard (predecision) discounted state-value function is defined as~\citep{Sutton2018}
\begin{equation}
V_\pi(s)\coloneqq\mathbb{E}_\pi\!\left[\sum_{j=0}^{\infty}\gamma^j\ell\bigl(s_j,\pi(s_j)\bigr)\,\middle|\,s_0=s\right],
\label{eq:gp-value-function}
\end{equation}
assuming finite expected discounted cost. MC value fitting estimates this expectation conditional on the current state. Nevertheless, evaluating it at a nominal terminal state neglects the uncertainty in the stochastic transition leading to that state. Accounting
for this uncertainty requires an additional expectation over possible terminal states, which can be approximated through explicit uncertainty propagation~\citep{Landgraf2026}.

To clarify the distinction, consider the final transition from $k+H_p-1$ to $k+H_p$, with all preceding horizon predictions treated nominally. Under the sufficiency condition in Eq.~\eqref{eq:postdecision-sufficiency}, Eqs.~\eqref{eq:postdecision-transition} and~\eqref{eq:postdecision-value} give
\begin{equation}
\mu_{\mathrm{term}}\gamma^{H_p}
\mathbb{E}\!\left[
V_\pi\!\left(F_{\mathrm w}(s_{k+H_p-1\mid k}^p,w_{k+H_p-1})\right)
\,\middle|\,s_{k+H_p-1\mid k}^p
\right]
=\mu_{\mathrm{term}}\gamma^{H_p}
V_\pi^p(s_{k+H_p-1\mid k}^p).
\label{eq:terminal-expectation-equivalence}
\end{equation}
Here, $V_\pi$ accounts for costs from $k+H_p$ onward conditional on the realized successor state, while the outer expectation averages over the final innovation that generates that state. The postdecision state-value function $V_\pi^p$ incorporates both sources of stochastic expectation, allowing direct nominal terminal evaluation without an additional online expectation. In contrast, evaluating $V_\pi$ at the nominal predecision terminal state $\bar s_{k+H_p\mid k}$ generally neglects the expectation over the final stochastic transition.

The training pairs in Eq.~\eqref{eq:postdecision-training-pair} associate nominal postdecision state features with MC returns beginning at the corresponding disturbed successor states. Regression therefore approximates the final-transition expectation together with the subsequent expected continuation cost. With the implemented features and finite-rollout targets, this is an approximation to Eq.~\eqref{eq:terminal-expectation-equivalence}, with target damping applied through Eq.~\eqref{eq:value-space-soft-update}. The GP posterior mean estimates the expected continuation cost, while the uncertainty-regulated terminal function in Eq.~\eqref{eq:risk-terminal} additionally accounts for critic epistemic uncertainty. Predictions within the MPC horizon remain certainty-equivalent (CE), as specified in Eq.~\eqref{eq:gprlmpc-objective}.

In practical implementation, training rollouts can be generated using a simulator or world model to reduce experimental costs. Plant data can then be used for model calibration, validation, and subsequent refinement. The learned critic approximates expected returns under the simulated dynamics and disturbance distribution. Increasing the number of MC rollouts reduces sampling error but cannot eliminate bias arising from model-plant mismatch. Reliable deployment consequently requires validation over the intended operating region and, when necessary, iterative refinement of the dynamics and disturbance models using plant data. Importantly, the GP posterior uncertainty characterizes uncertainty in the learned critic conditional on the assumed model and training data; it does not directly quantify simulator-plant discrepancy. Robustness to such mismatch must therefore be assessed separately.

\section{Postdecision State Data, Targets, and Terminal Evaluation}
\label{app:postdecision-conventions}

Fix a rollout-generating policy $\pi$ and collect $R$ disturbed rollouts, each containing $T$ stage costs. Initial states, previous inputs, reference schedules, and controller and estimator initialization are matched across rollouts and policies. Disturbance seed pairs differ across rollouts and are reused across policies. For clarity, PI and rollout superscripts are omitted, and returns are computed separately within each rollout.

In the implementation, we define $s_{k+l\mid k}^p\coloneqq\bar s_{k+l+1\mid k}$ within prediction horizon $H_p$, where the overbar denotes a nominal prediction based on the predecision states, with disturbance innovations set to zero. Thus, $s_{k+l\mid k}^p$ is indexed by the generating action at $k+l$ and represents the nominal endpoint at $k+l+1$. Let $\varphi$ denote the critic feature map and define $\chi_k^p=\varphi(s_{k\mid k}^p)=\varphi(\bar s_{k+1\mid k})$. Pair this nominal feature with the cost-to-go from the actual disturbed trajectory:
\begin{equation}
\left(\chi_k^p,G_{k+1}\right),\qquad
G_{k+1}\coloneqq\sum_{j=k+1}^{T-1}\gamma^{j-k-1}\ell(s_j,u_j).
\label{eq:postdecision-training-pair}
\end{equation}
Returns are computed backward using
\begin{equation}
G_T=0,\qquad
G_j=\ell(s_j,u_j)+\gamma G_{j+1}.
\label{eq:mc-postdecision-returns}
\end{equation}
Each target excludes the current stage cost $\ell(s_k,u_k)$ and contains no additional leading discount. The first training pair is $(\chi_0^p,G_1)$, so $G_0$ is not used as a target and the initial stage cost $\ell(s_0,u_0)$ is excluded from the continuation targets. The boundary pair $(\chi_{T-1}^p,G_T)$ is also omitted because its zero target is imposed by finite-rollout truncation. To reduce truncation bias, discard the final $N_{\mathrm{eff}}$ remaining pairs within each rollout. For $0<\gamma<1$, define
\begin{equation}
N_{\mathrm{eff}}=\left\lceil\frac{\log(1-\zeta)}{\log\gamma}\right\rceil,\qquad 0<\zeta<1.
\label{eq:effective-return-horizon}
\end{equation}
Here, $\zeta$ specifies discount-weight coverage, with $\gamma^{N_{\mathrm{eff}}}\leq1-\zeta$. The retained training pairs correspond to action indices from $k=0$ through $k=T-2-N_{\mathrm{eff}}$ within each rollout. For $T>N_{\mathrm{eff}}+1$, concatenate the retained pairs from all rollouts and reindex them as
\begin{equation}
\mathcal D^p=\left\{(\chi_i^p,G_i)\right\}_{i=1}^{N_{\mathrm{pool}}},
\qquad
N_{\mathrm{pool}}=R(T-1-N_{\mathrm{eff}}).
\label{eq:pooled-postdecision-data}
\end{equation}
Here, $i$ indexes pooled pairs rather than time, each $G_i$ denotes the corresponding continuation return $G_{k+1}$ from its source rollout, and repeated samples are retained. Each critic fit uses the current policy-specific batch. The dataset stores raw returns; soft targets and parameter warm starts are applied as described in Section~\ref{ssec:soft-update}.

Using the uncertainty-regulated terminal estimate $V_\theta^p$ defined in Eq.~\eqref{eq:risk-terminal}, the terminal query and its objective contribution are
\begin{equation}
\chi_{k+H_p-1\mid k}^p=\varphi(\bar s_{k+H_p\mid k}),
\qquad
\mu_{\mathrm{term}}\gamma^{H_p}
V_\theta^p\bigl(\chi_{k+H_p-1\mid k}^p\bigr).
\label{eq:postdecision-terminal-convention}
\end{equation}
The horizon sum includes stage costs from $k$ through $k+H_p-1$, while the continuation term is aligned with costs beginning at $k+H_p$. GPR can estimate conditional return means without requiring identical feature vectors to recur. The fitted GP mean therefore provides a feature-based approximation under the sampled rollout distribution, with target damping applied during fitting.

\section{Experiment Configurations}
\label{app:experiment-configurations}

\begin{table}[!t]
    \centering
    \footnotesize
    \setlength{\tabcolsep}{3pt}
    \caption{Plant, controller, and operating settings for the CSTR experiments.}
    \label{app-tab:cstr-parameters}
    \begin{tabular}{@{}
        p{0.23\linewidth}
        p{0.13\linewidth}
        p{\dimexpr0.64\linewidth-4\tabcolsep\relax}
    @{}}
        \toprule
        Parameter & Value & Description \\
        \midrule
        \multicolumn{3}{@{}l}{\textit{Plant parameters}} \\
        $k_1$ & $50.0$ & Rate constant for $A\rightarrow B$ [$\mathrm{h}^{-1}$] \\
        $k_2$ & $100.0$ & Rate constant for $B\rightarrow C$ [$\mathrm{h}^{-1}$] \\
        $k_3$ & $10.0$ & Rate constant for $2A\rightarrow D$ [$\mathrm{L\,mol^{-1}\,h^{-1}}$] \\
        $x_{1f,\mathrm{nom}}$ & $10.0$ & Nominal feed concentration [$\mathrm{mol\,L^{-1}}$] \\
        $\sigma_{x_2}$ & $2.0$ & Product-concentration forcing-rate std. [$\mathrm{mol\,L^{-1}\,h^{-1}}$] \\
        $\sigma_{x_{1f}}$ & $10.0$ & Feed-concentration forcing-rate std. [$\mathrm{mol\,L^{-1}\,h^{-1}}$] \\
        \midrule
        \multicolumn{3}{@{}l}{\textit{Initial conditions and constraints}} \\
        $x_{1,0}$ & $2.15$ & Initial concentration of $A$ [$\mathrm{mol\,L^{-1}}$] \\
        $x_{2,0}$ & $0.90$ & Initial concentration of $B$ [$\mathrm{mol\,L^{-1}}$] \\
        $x_{1f,0}$ & $10.00$ & Initial feed concentration [$\mathrm{mol\,L^{-1}}$] \\
        $u_{-1}$ & $19.63$ & Input preceding the first control move [$\mathrm{h}^{-1}$] \\
        $x_{1,\min}$ & $0.0$ & Lower bound on $x_1$ [$\mathrm{mol\,L^{-1}}$] \\
        $x_{1,\max}$ & $7.0$ & Upper bound on $x_1$ [$\mathrm{mol\,L^{-1}}$] \\
        $x_{2,\min}$ & $0.0$ & Lower bound on $x_2$ [$\mathrm{mol\,L^{-1}}$] \\
        $x_{2,\max}$ & $2.4$ & Upper bound on $x_2$ [$\mathrm{mol\,L^{-1}}$] \\
        $u_{\min}$ & $0.0$ & Lower dilution-rate bound [$\mathrm{h}^{-1}$] \\
        $u_{\max}$ & $60.0$ & Upper dilution-rate bound [$\mathrm{h}^{-1}$] \\
        \midrule
        \multicolumn{3}{@{}l}{\textit{Controller and simulation parameters}} \\
        $H_p$ & $3$ & Prediction horizon [steps] \\
        $T$ & $1000$ & Rollout length [sampling steps] \\
        $T_s$ & $0.005$ & Sampling interval [$\mathrm{h}$] \\
        $T_{\mathrm{int}}$ & $1.0$ & Reference update interval [$\mathrm{h}$] \\
        $\gamma$ & $0.98$ & Discount factor \\
        $(\mu_y,\mu_u,\mu_{\mathrm{term}})$ & $(10^2,1,1)$ & Tracking, input-move, and terminal weights \\
        $\zeta$ & $0.99$ & Effective-horizon return coverage factor \\
        $r_{\min}$ & $0.7$ & Lower bound on reference [$\mathrm{mol\,L^{-1}}$] \\
        $r_{\max}$ & $0.9$ & Upper bound on reference [$\mathrm{mol\,L^{-1}}$] \\
        $r_0$ & $0.9$ & Initial reference value [$\mathrm{mol\,L^{-1}}$] \\
        $R$ & $5$ & Number of rollouts per policy \\
        $M_{\max}$ & $80$ & Maximum number of PC-selected inducing points \\
        \bottomrule
    \end{tabular}
\end{table}

Table~\ref{app-tab:cstr-parameters} summarizes the plant, controller, and operating settings. Both controllers are implemented using CasADi with IPOPT. No separate terminal set is imposed beyond the prescribed state bounds. The disturbed plant and nominal one-step predictor use CVODES with absolute and relative tolerances of $10^{-10}$. Full augmented-state feedback is assumed without additional measurement noise. The reference is initially $0.9\,\mathrm{mol\,L^{-1}}$ and subsequently sampled uniformly from $[0.7,0.9]\,\mathrm{mol\,L^{-1}}$ at each reference update; its known future values are supplied to controllers. Reference schedules and disturbance seed pairs are reused across policies, with all rollout initializations reset as described in Appendix~\ref{app:postdecision-conventions}.

Critic data are split into training and diagnostic subsets in an $80\%{:}20\%$ ratio using ten target-quantile strata. We apply $\tau=0.1$ for soft updates, fixed first-round scalers, and parameter warm starts. The mean is $m_\beta(\chi)=\beta_2\chi_{\mathrm{err}}^2+\beta_1\chi_{\mathrm{err}}+\beta_0$, where $\chi_{\mathrm{err}}$ is the standardized tracking-error feature. Adam uses a learning rate of $10^{-3}$ for at most $5000$ steps, stopping after $200$ consecutive steps without a training-loss improvement exceeding $10^{-4}$. PC uses a remaining-trace tolerance of $10^{-3}$ and a residual-variance cutoff of $10^{-12}$; DTC uses $\delta_{\mathrm{jit}}=10^{-8}$ in standardized coordinates. Screened runs terminate at the first SSE deterioration or upon exhausting the prescribed PI budget; rejected candidates are excluded from the accepted sequence. Diagnostic runs use the matched iteration budgets described in Section~\ref{ssec:risk-coefficient}.

\section{Closed-Loop Trajectories}
\label{app:cl-results}

\begin{figure}[!htbp]
    \centering
    \includegraphics[width=\linewidth]{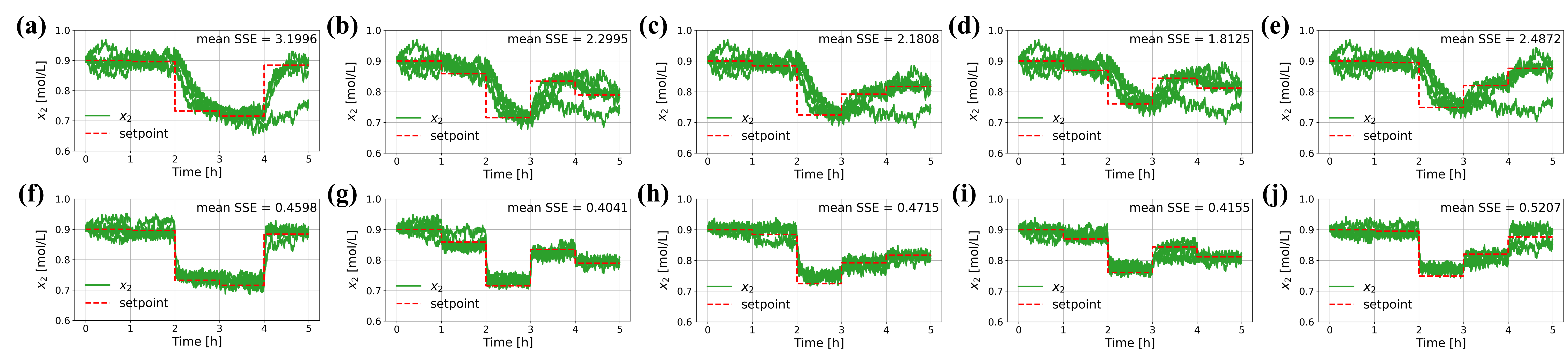}
    \caption{Multi-seed closed-loop tracking responses: (a)--(e): Baseline MPC; (f)--(j): Final incumbent GP--RL--MPC controllers with $\kappa=3.0$. Each panel overlays $R$ rollouts.}
    \label{fig:mpc-comparison}
\end{figure}

Fig.~\ref{fig:mpc-comparison} compares the five-seed closed-loop trajectories under the matched settings described in Section~\ref{ssec:baseline-pi-performance}. GP--RL--MPC tracks the varying setpoints more closely and reduces persistent offsets relative to baseline MPC.

\end{document}